%% file: paper_arxiv.tex
\documentclass[journal]{IEEEtran}
\usepackage{hyperref}       
\usepackage{url}            
\usepackage{booktabs}       
\usepackage{amsfonts}       
\usepackage{nicefrac}       
\usepackage{microtype}      
\usepackage{graphicx}
\usepackage{floatrow}
\usepackage{bm}
\usepackage{amsmath}
\usepackage{color}
\usepackage{floatrow}
\usepackage{cite}
\usepackage{soul}
\usepackage{amsthm}
\usepackage{comment}
\usepackage{float}
\usepackage{tablefootnote}
\usepackage[ruled,linesnumbered]{algorithm2e}
\usepackage{threeparttable}
\usepackage{verbatim} 
\usepackage{bbding}
\newcommand{\xqedhere}[2]{%
	\rlap{\hbox to#1{\hfil\llap{\ensuremath{#2}}}}}

\soulregister\cite7
\soulregister\ref7
\soulregister\eqref7

\ifCLASSINFOpdf
\else
\fi
\begin{document}
%
\title{Self-Supervised Point Cloud Registration with Deep Versatile Descriptors for Intelligent Driving}

\author{Dongrui Liu, 
        Chuanchaun Chen, 
        Changqing Xu,
        Robert C. Qiu,
         and  Lei Chu 
\thanks{All authors are with Shanghai Key Laboratory of Navigation and Location based Services, School of Electronic Information and Electrical Engineering, Shanghai Jiao Tong University.}}
\markboth{}%
{Shell \MakeLowercase{\textit{et al.}}: Bare Demo of IEEEtran.cls for IEEE Journals}

\maketitle

\begin{abstract}
As a fundamental yet challenging problem in intelligent transportation systems, point cloud registration attracts vast attention and has been attained with various deep learning-based algorithms. The unsupervised registration algorithms take advantage of deep neural network-enabled novel representation learning while requiring no human annotations, making them applicable to industrial applications. However, unsupervised methods mainly depend on global descriptors, which ignore the high-level representations of local geometries. In this paper, we propose to jointly use both global and local descriptors to register point clouds in a self-supervised manner, which is motivated by a critical observation that all local geometries of point clouds are transformed consistently under the same transformation. Therefore, local geometries can be employed to enhance the representation ability of the feature extraction module. Moreover, the proposed local descriptor is flexible and can be integrated into most existing registration methods and improve their performance. Besides, we also utilize point cloud reconstruction and normal estimation to enhance the transformation awareness of global and local descriptors. Lastly, extensive experimental results on one synthetic and three real-world datasets demonstrate that our method outperforms existing state-of-art unsupervised registration methods and even surpasses supervised ones in some cases. Robustness and computational efficiency evaluations also indicate that the proposed method applies to intelligent vehicles.
\end{abstract}


\IEEEpeerreviewmaketitle

\section{INTRODUCTION}

\IEEEPARstart{P}{oint} clouds are one of the vital industrial measurements in intelligent transportation systems. The corresponding registration is a critical yet demanding task with various applications in many fields, including large-scale reconstruction \cite{agarwal2011building}, mapping \cite{li2018collaborative, ye2020robust}, and object detection and localization \cite{wen2019gnss, liu2018dynamic, wong2017segicp}. For autonomous driving, point cloud registration utilizes local scans of the same object from different recording angles to provide complete and high-quality 3D representations. Besides, point cloud registration can match a real-time 3D view to an existing 3D map to provide localization results \cite{huang2021comprehensive}. In this way, autonomous vehicles can estimate their positions on the map and adjust real-time route planning.

\par The last two decades have witnessed the rapid development of point cloud registration, aiming to align one point cloud (source) to another (target) by finding the optimal rigid transformation. Iterative Closest Point (ICP) \cite{besl1992method} is a widely used method with two iterative optimization processes: correspondences searching and rigid transformation estimation. Nevertheless, ICP is sensitive to initialization and easily deceived by local optima. Many methods \cite{chetverikov2002trimmed, fitzgibbon2003robust, segal2009generalized, bouaziz2013sparse, yang2015go} have been proposed to tackle these issues with increased computational complexity. On the other hand, with the rapid development of deep learning techniques, the rigid point cloud registration problem has been revisited with many novel designs. With ground-truth correspondences or category labels, recent studies \cite{aoki2019pointnetlk, wang2019deep, yew2020rpm, li2020iterative, yuan2020deepgmr, li2021pointnetlk} tried to solve the problem in a supervised manner. These methods effectively obtain the informative feature of point cloud and achieve superior performances, requiring high quality and quantity of ground-truth correspondences. However, there may not exist point-point correspondences in real-world intelligent transportation systems in the presence of noises and outliers. 
\begin{figure}[t]
	\centering
	\includegraphics[width=0.95\columnwidth]{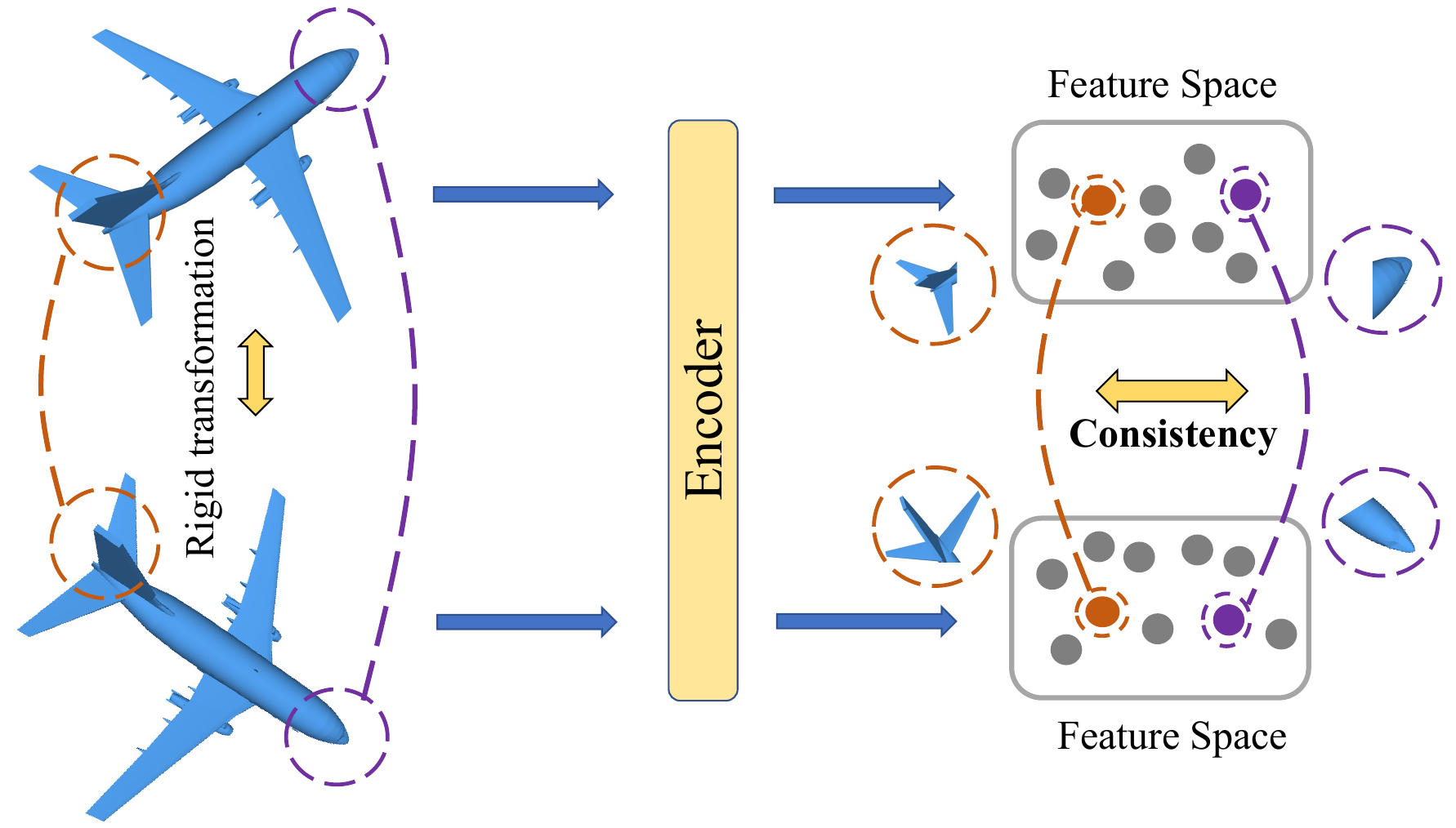} 
	\caption{\textbf{ The illustration of the geometrical consistency among distinctive local subsets in a point cloud pair.} Given a \textit{source} point cloud and a \textit{target} point cloud, under the rigid transformation, all local geometries of point clouds are transformed consistently with the same transformation. In this paper, we select representative and distinctive local subsets to ensure no abuse of all local geometries.}
	\label{fig1}
	\vspace{-0.4cm}
\end{figure}

Realizing registration from unlabeled point cloud data without ground-truth correspondences is a significant yet rarely explored challenge in the literature. To our best knowledge, there exist only two works \cite{sarode2019pcrnet,huang2020feature} so far. PCRNet \cite{sarode2019pcrnet} alleviates the pose misalignment shown in the PointNetLK \cite{aoki2019pointnetlk} by replacing the Lucas-Kanade module with multi-layered perceptrons. PCRNet directly recovers transformation parameters from concatenated global descriptors of source and target point cloud. The other significant work, FMR-Net \cite{huang2020feature}, employs an autoencoder framework with the PointNetLK being its backbone and then achieves the registration by minimizing the feature-metric projection error. Although these two approaches showcase the feasibility of applying unsupervised learning, they mainly depend on global descriptors. Global descriptors are insufficient to represent the whole geometry of a point cloud and are challenging to encode local geometries, especially when handling partial point clouds. Thus, it is necessary to figure out a more robust way to encode global and local geometries.

In this paper, we jointly utilize global and local descriptors in a self-supervised manner to enhance deep unsupervised point cloud registration. We are motivated by wide applications of self-supervised learning on visual tasks \cite{komodakis2018unsupervised, chen2021exploring, senanayaka2020toward} in recent years. As a branch of unsupervised learning, self-supervised learning frameworks usually design some pretext tasks, and train convolutional neural networks (CNNs) for solving these pretext tasks, \emph{e.g.}, predicting image rotation \cite{komodakis2018unsupervised} or classifying positive and negative pairs \cite{chen2021exploring}. These pretext tasks enable CNNs to extract high-level semantic descriptors. Therefore, we aim to extract local and high-level global descriptors, which is beneficial for solving registration through a pretext task. Specifically, Fig. \ref{fig1} shows a critical observation that the rigid transformation consistently transforms different local regions and preserves global and regional structural information. Based on such an observation, we design a pretext task to encode global and local descriptors and propose a local consistency loss to ensure the intrinsic geometric consistency among local distinctive subset points, enhancing the representation ability of the feature extraction module.

Furthermore, unsupervised point cloud registration techniques align two point clouds, mainly relying on differences between two global descriptors. However, they ignore a crucial issue of obtaining global descriptors with high transformation awareness, \emph{i.e.,} differences between global descriptors of two-point clouds should be significant when the two-point clouds are not well aligned. To tackle this problem, we present two additional tasks: self-reconstruction and normal estimation. In summary, our main contributions can be summarized as follows:
\begin{itemize}
	\item To the best of our knowledge, this paper is the first to formulate the registration problem by both global and local descriptors in a self-supervised manner, requiring no labeled data or arduously searching correspondences. 
	
	\item  The proposed local descriptor is a simple yet effective module that applies to most existing methods and improves the performances of both correspondence-based and correspondence-free registration methods.

	\item Comprehensive experiments have been conducted on both synthetic and real-world datasets, demonstrating that the proposed method outperforms the existing unsupervised ones and surpasses supervised ones in some cases.
	
\end{itemize}

 The remainder of this paper is organized as follows. Section II introduces related work. Problem formulation is shown in Section III. Section IV elaborates on the proposed method and related discussions. Experiments are given in Section V. Section VI concludes this work.

\section{Related Work}
Point cloud registration is a vital research field due to various 3D perception applications. Here we broadly categorize the related work into \textit{traditional}, \textit{supervised}, and \textit{unsupervised} registration methods.

\subsection{Traditional Registration}
ICP \cite{besl1992method} is the well-known method for the rigid point cloud registration, which iteratively computes the closet point as correspondences and estimates the optimal transformation by the correspondences. Subsequent works have been proposed to improve the robustness of ICP \cite{segal2009generalized,bouaziz2013sparse,chetverikov2002trimmed}. On the other branch, Gaussian Mixture Model (GMM) \cite{jian2010robust} and Hierarchical Gaussian Mixture Registration (HGMR)\cite{eckart2018hgmr} reformulate the registration problem as probability distribution matching. These probabilistic methods still require good initializations to avoid being trapped into the local optimum, because their objective functions are non-convex. Unlike local registration methods, global registration methods do not require good initializations. In this way, Go-ICP \cite{yang2015go}, GOGMA \cite{campbell2016gogma}, GOSMA \cite{campbell2019alignment}, and GoTS \cite{liu2018efficient} have been proposed to find globally optimal registration by branch-and-bound (BnB) method at the expense of increased computational complexity. In another line, PFH \cite{rusu2008aligning} and FPFH \cite{rusu2009fast} constructed handcrafted feature descriptors and estimated potential correspondences. Then robust optimization approaches such as semidefinite programming \cite{yang2020teaser} and Random Sampling Consensus (RANSAC) \cite{fischler1981random} can be utilized to estimate exact correspondences. Fast Global Registration (FGR) \cite{zhou2016fast} accelerated the optimization process by a graduated non-convex strategy. However, requiring good initializations, noisy correspondences, and time constraints are challenging for traditional registration techniques.

\subsection{Deep Supervised Registration}
The rapid development of deep learning methods paves the way for developing a new perspective for point cloud registration \cite{qi2017pointnet, 10091154, liu2022pfmixer}. Recently, PointNetLK \cite{aoki2019pointnetlk} utilizes the PointNet module to obtain the global descriptors and aligns two point clouds iteratively by minimizing the distances between two learned descriptors with a modified Lucas-Kanade (LK) algorithm \cite{baker2004lucas}. Deep Closest Point \cite{wang2019deep} introduces a solution in another perspective that extracts per-point features and uses a transformer followed by a differentiable Singular Value Decomposition (SVD) module to recover the transformation. PRNet \cite{wang2019prnet} uses a keypoint detector to establish keypoint correspondences to solve the partial to partial point cloud registration in a self-supervised way. RPMNet \cite{yew2020rpm} proposes Sinkhorn normalization to estimate soft correspondences. IDAM \cite{li2020iterative} computes pairwise correspondences by an iterative distance-aware similarity convolution module. DeepGMR \cite{yuan2020deepgmr} learns point-to-GMM correspondences by integrating GMM. PBNet \cite{zheng2022global} proposes a two-stage registration process, including a deep learning stage and a traditional optimal search stage. In addition, many representative techniques\cite{deng20193d, choy2019fully, choy2020deep, li2021point, ao2021spinnet} have been presented to deal with large-scale point cloud registration. For instance, GeoTransformer \cite{qin2022geometric} uses the well-established transformer structure for point cloud registration and achieves promising results. While deep supervised registration methods achieve state-of-the-art performances, it is worth exploring unsupervised ones that require no human annotations, offering the demanding flexibility and applicability of registration algorithm design in realistic scenes, \emph{i.e.}, intelligent driving.

\subsection{Unsupervised Registration}
In contrast to the supervised and correspondence-based point cloud registration methods, there are two prior attempts at unsupervised point cloud registration. PCRNet \cite{sarode2019pcrnet} improves PointNetLK by replacing the LK module with multi-layered perceptrons. FMR-Net \cite{huang2020feature} adapts PointNetLK to a semi-supervised manner by jointly optimizing the processes of feature extraction and transformation estimation. However, they achieve registration only depending on global representations, paying no attention to the valuable representations obtained by local geometries. In this way, the representation ability of neural networks is restrained, and it is challenging to characterize the geometries of point clouds.

Meanwhile, as a subclass of unsupervised learning, self-supervised learning has achieved remarkable performance on visual tasks without expensive labels \cite{senanayaka2020toward, achituve2021self, rao2020global} by training CNNs for solving some pretext tasks. These pretext tasks enable CNNs to extract high-level semantic descriptors. 
As a result, beyond the global descriptors, we will delve into the local ones in a self-supervised manner for enhanced deep unsupervised registration.

\section{Problem Formulation}
\label{secIII}
In this section, we present the formulation of the rigid registration problem. We denote ${X} =\lbrace {\bm x}_{1},{ \bm x}_{2}, \ldots, {\bm x}_{N} \rbrace$ $\subset \mathbb R^{3}$ and ${Y} =\lbrace {\bm y}_{1},{ \bm y}_{2}, \ldots, {\bm y}_{M} \rbrace$ $\subset \mathbb R^{3}$ as the \textit{source} and the \textit{target} point cloud, respectively, where $N$ and $M$ represent the point number in source and target point cloud, respectively. The objective of registration is to estimate the rigid transformation $\lbrace\bm{ R, t} \rbrace$ as \cite{yang2015go}:

\begin{equation}
\bm {R, t} = \mathop{\arg \min}_{\bm R\in SO(3), \bm t\in \mathbb R^{3}} \sum_{i=1}^{N}\| \bm {R}{\bm x}_{i}+\bm t -{\bm y}_{j*}\|_{2},
\label{1}
\end{equation}
where ${\bm R}\in SO(3)$ represents a rotation matrix and ${ \bm t \in \mathbb R^{3} }$ denotes a translation vector. Given ${\bm R}$ and $\bm t $, the optimal correspondence of ${\bm x}_{i}$ can be searched as follows.
\begin{equation}
j^{*}=\underset{j \in\{1, \ldots, M\}}{\operatorname{argmin}}\left\|\bm{R} \bm{x}_{i}+\bm{t}-\bm{y}_{j}\right\|_2.
\label{corrspondence}
\end{equation}
Let the point ${\bm y}_{j*} \in Y$ denote the optimal correspondence of ${\bm x}_{i}$, \emph{i.e.} ${\bm y}_{j*}$ is the closest point to the point ${\bm R}{\bm x}_{i}+\bm t$ in $Y$.
Eq. \eqref{1} is a non-convex problem and cannot be trivially solved. If true correspondences between $X$ and $Y$ are known, transformation parameters can be optimally estimated in closed-form \cite{horn1987closed, arun1987least}. If optimal
transformation parameters are given, correspondences can be easily searched. ICP solves the problem by alternatively searching correspondences and estimating transformation parameters. However, such an iterative optimization process guarantees convergence to a local minimum \cite{besl1992method}.

To address the issue, PointNetLK \cite{aoki2019pointnetlk} aims to estimate the optimal transformation in the feature space instead of Euclidean space. It utilizes a neural network, namely PointNet \cite{qi2017pointnet}, to obtain a $K$-dimensional vector as global representations of point clouds. In this way, the registration problem can be described as follows:
\begin{equation}
\bm {R, t}=\underset{\bm R\in SO(3), \bm t\in \mathbb R^{3}}{\arg \min }\|\psi(\bm {R}X+\bm t)-\psi(Y)\|_{2},
\label{2}
\end{equation}
where $\psi: \mathbb{R}^{3 \times N} \rightarrow \mathbb{R}^{K}$ denotes the feature extraction functional. According to Eq. (\ref{2}), transformation parameters can be estimated without searching correspondences. Previous studies \cite{huang2020feature,aoki2019pointnetlk} primarily solved the Eq. \eqref{2} by adapting PointNet as the feature extraction function,  resulting in a global descriptor. However, they fail to utilize the informative features from local geometries. 

To address the issue, we reformulate point cloud registration by jointly considering global and local descriptors. Inspired by the success of self-supervised learning \cite{jing2020self, lan2019albert}, we design a pretext task to extract informative global and local descriptors. We discover a key observation that \textit{source} point cloud is transformed into \textit{target} point cloud by a rigid transformation operator $W$. In other words, all local geometries of point clouds are transformed consistently with the same mapping function $W$. This property inspires us to exploit the high-level relations among different local geometries. 

Given two subsets ${\left\{ {{X_k}} \right\}_{k = 1,2}} \subset X$ and the corresponding transformed version ${\left\{ {W({X_k})} \right\}_{k = 1,2}}$, the related high-level representations can be denoted by ${\left\{ {\psi ({X_k})} \right\}_{k = 1,2}}$, and ${\left\{ {\psi ({W(X_k)})} \right\}_{k = 1,2}}$, respectively. $\psi$ represents a feature extraction function. Accordingly, we have local descriptors $\psi(X_{k})$ and $\psi(W(X_{k}))$. In this way, the feature change of original and transformed $X_{k}$ is described as 
\begin{equation}
\tau(X_{k}) \stackrel{\text {def}}{=}\mathcal M[\psi(X_{k}), \psi(W(X_{k}))], 
\label{3}
\end{equation}
where $\mathcal M$ is a metric function for modeling differences of features. We employ a Fully Connected (FC) layer $FC:\mathbb{R}^{2K} \rightarrow \mathbb{R}^{K}$ as the metric function $\mathcal M$, such that
\begin{equation}
\mathcal M[\psi(X_{k}),\psi(W(X_{k}))] = FC[\psi(X_{k})\oplus\psi(W(X_{k}))],
\label{FEF1}
\end{equation}
where $\oplus$ is the concatenation operation and $FC:\mathbb{R}^{2K} \rightarrow \mathbb{R}^{K}$. In Eq. \eqref{FEF1}, high-dimensional local descriptors $\psi(X_{k})$ and $\psi(W(X_{k}))$ are combined together with $\mathcal M$, whose goal is to perform feature extraction while maintaining the change of each feature.\footnote{The employment of the FC layer is motivated by the network design in PointNet \cite{qi2017pointnet}, and FC network-based intra prediction for image coding \cite{li2018fully}.}
The goal of $\tau$ is to abstract high-level relations between the local region and its transformed version. 

\begin{figure}[t]
	\centering
	\includegraphics[width=0.9\columnwidth]{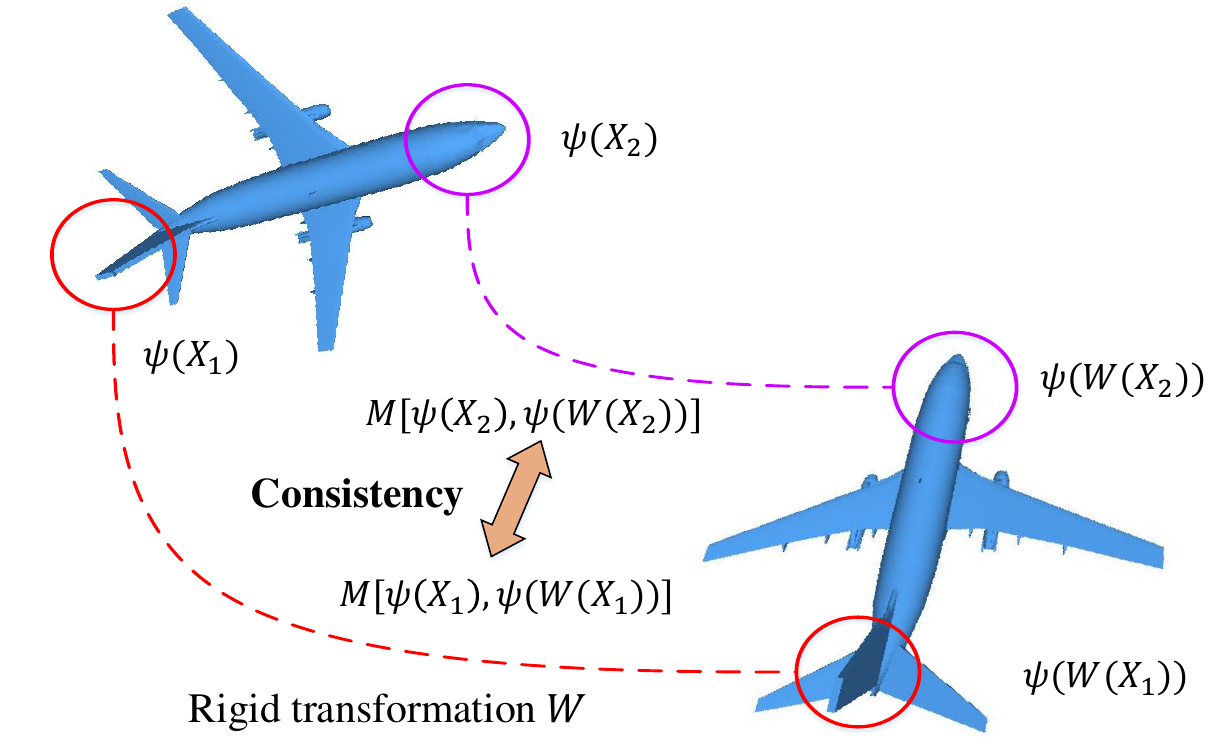}
	\caption{\textbf{The illustration of the key idea in the proposed registration method.}  We use a pretext task to force the encoder to extract versatile descriptors. We model the relation of different local geometries under a rigid transformation because various local parts of point clouds are transformed consistently. Specifically, given a head (purple circle) and a tail (red circle), changes in transformed and original descriptors of these two different local regions should be consistent.}
	\label{motivation}
	\vspace{-0.3cm}
\end{figure}

Since all local geometries of point clouds are transformed consistently with the same mapping function $W$, the feature change between different regions should be consistent, \emph{i.e.}, the distributions of $\tau(X_{1})$ and $\tau(X_{2}) $ are same, as shown in Fig. \ref{motivation}. In this way, we have $\tau(X_{1}) \sim \mathbb Q$ and $\tau(X_{2}) \sim \mathbb Q$, where $\mathbb Q$ denotes a distribution. Based on the above analysis, the registration problem in Eq. \eqref{2} can be reformulated in a self-supervised manner as
\begin{equation}\begin{array}{ll}
&\underset{W \in SE(3), \psi}{\arg \min }\|\psi(W(X))-\psi(Y)\|_{2}, \\
&\text { subject to } \tau(X_{1}) \sim \mathbb Q, \tau(X_{2}) \sim \mathbb Q. \\
\end{array}
\label{313}
\end{equation}
Here, we are trying to deal with the point cloud registration problem with a joint design on global and local descriptors, which are respectively shown in the objective and the constraint of Eq. \eqref{313}.

However, the constraint $\tau(X_{1}) \sim \mathbb Q, \tau(X_{2}) \sim \mathbb Q$ in Eq. \ref{313} is non-convex. This constraint means $\tau(X_{1}) $ and $\tau(X_{2}) $ should be sampled from the same distribution, which is extremely hard to achieve for a network functional. Alternatively, we try to tackle the problem by minimizing the distance metric $D$ of their distributions as 
\begin{equation}
\underset{W \in SE(3), \psi}{\arg \min } \|\psi(W(X))-\psi(Y)\|_{2} + D(\tau(X_{1}) \| \tau(X_{2})).
\label{4}
\end{equation}

In this way, we need to train a feature extraction function $\psi$ and estimate a transformation operator $W$. Given that the optimization of the function $\psi$ and the transformation operator $W$ in Eq. \eqref{4} are difficult and time-consuming, we employ a
simplified training strategy to optimize $\psi$ and $W$ in an alternative manner. \emph{I.e.}, when the transformation operator $W$ is optimized and the function $\psi$ is frozen in each iteration and vice versa. Specifically, we will introduce how to train the function $\psi$ and estimate the optimal $W$ in the next section.

\begin{figure*}[t]
	\centering
	\includegraphics[width=0.9\textwidth]{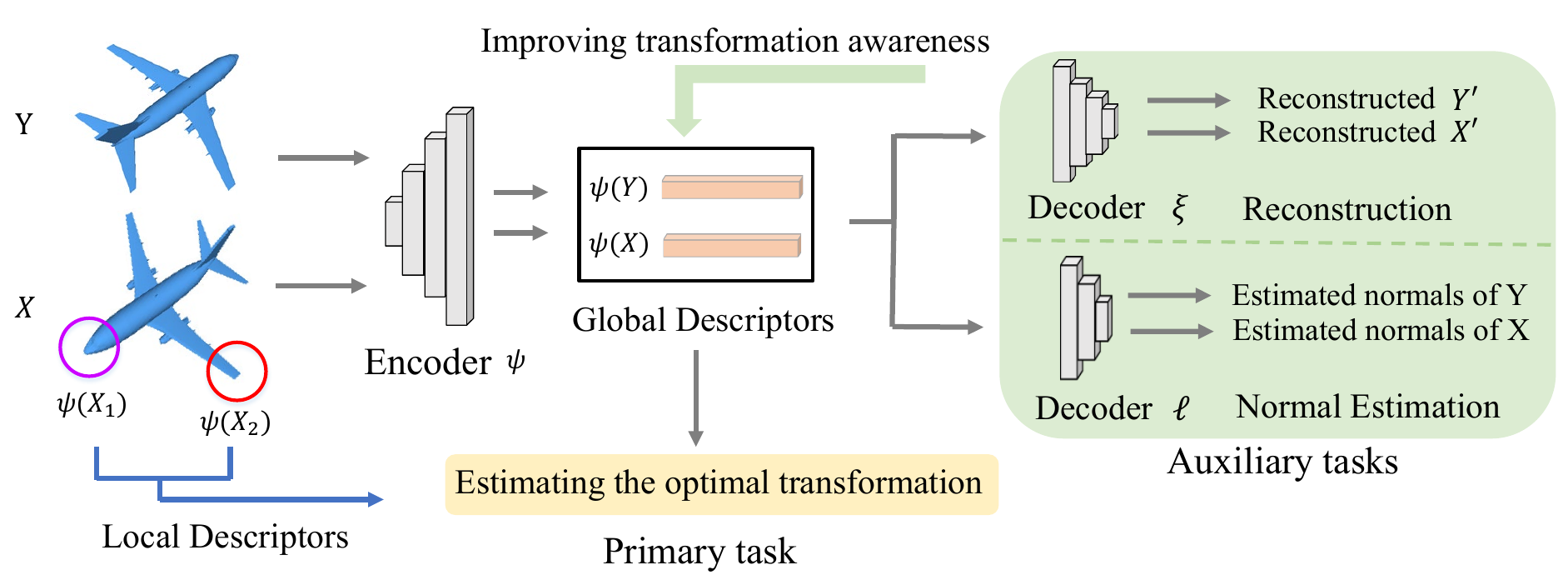} 
	\caption{The overall architecture of the proposed unsupervised point cloud registration method. The encoder learns global and local descriptors of source and target point clouds ($X$ and $Y$). The primary task is to solve point cloud registration jointly based on global and local descriptors, $\psi(X) $ and $\psi(X_k),\ k\in{\left\{1,2 \right\}}$, in a self-supervised manner. We further introduce two auxiliary tasks to enhance the transformation awareness of global and local descriptors. }
	\label{fig2}
\end{figure*}

\section{Robust Point Registration}

The core of rigid-body registration without correspondences is to extract discriminative and transformation awareness features from two-point clouds and recover the related transformation parameters. We solve the problem by leveraging the proposed versatile descriptors with two auxiliary tasks. The overall framework is illustrated in Fig. \ref{fig2}. In short, we first embed the point clouds in high-dimensional space, performing optimization with one primary task ${{{\cal L}_{{\rm{p}}}}}$ and two auxiliary tasks (${{{\cal L}_{{\rm{CD}}}}}$ and ${{{\cal L}_{{\rm{n}}}}}$). The primary task estimates the optimal transformation parameters while considering both global and local descriptors in a self-supervised way. Auxiliary tasks aim to improve the transformation awareness of the feature extraction function.

The overall objective $\mathcal{L}$ to be minimized in the proposed method can be formalized as
\begin{equation}
{\cal L} = \underbrace {{{\cal L}_{{\rm{p}}}}}_{\rm Primary \ task} + \ {\lambda _1}  \underbrace {{{\cal L}_{{\rm{CD}}}}}_{\rm Auxiliary \ task \ I} + \ {\lambda _2} \underbrace {{{\cal L}_{{\rm{n}}}}}_{\rm Auxiliary \ task \ II},
\label{PI1}
\end{equation}
where trade-off parameters $\lambda_{1}$ and $\lambda_{2}$ are positive constants to balance the
effect of different tasks. The optimization of Eq. \eqref{PI1} takes the source and the target point cloud as inputs without requiring ground truth transformation, allowing us to learn unsupervised representations for robust point cloud registration.

\subsection{Primary Task: Learning with Local Consistency}
\label{LLC}
This subsection elaborates on the proposed scheme. Based on the problem formulation in Section \ref{secIII} and Eq. \eqref{4}, we try to tackle the unsupervised registration problem as 
\begin{equation}
\begin{aligned}
\mathcal{L}_{\rm p}= & \quad \|\psi(W(X))-\psi(Y)\|_{2} \quad + \\
&\frac{1}{2}(D_{K L}(\tau(X_{1}) \| \tau(X_{2})) + D_{K L}(\tau(X_{2}) \| \tau(X_{1}))).
\end{aligned}
\label{5}
\end{equation}
In Eq. \eqref{5}, the distribution alignment problem could also be measured by other distance metrics, such as Maximum Mean Discrepancy and Euclidean distance. We empirically employ the symmetric Kullback-Leibler divergence, which has stable and superior performance against other distance metrics.

The first term in Eq. \eqref{5} uses global descriptors to measure differences between the point cloud $X$ and the point cloud $Y$. The estimation of transformation parameters is based on the first term. However, estimating transformation $W$ is time-consuming. Therefore, we choose to use the computationally efficient Inverse Compositional (IC) method \cite{baker2004lucas} to iteratively calculate transformation increment $\Delta \bm w$ as 
\begin{equation}
\Delta \bm w=(J^{T}J)^{-1}J^{T} (\psi(W(X))-\psi(Y)),
\end{equation}
where $J$ is the Jacobian matrix of differences between global descriptors of source and target point clouds with respect to transformation parameters. Unlike regular grid images,
computing the Jacobian matrix of irregular point clouds needs to take gradients in $x$, $y$, and $z$ directions. Refer to \cite{aoki2019pointnetlk, li2021pointnetlk}, the warp Jacobian in the IC-LK algorithm can be approximated as
\begin{equation}
J=\frac{\partial \psi(W^{-1}(Y))}{\partial \bm w},
\end{equation}
where twist parameters $\bm w=(w_{1},w_{2}, \dots, w_{6})^{T} \in \mathbb R^{6}$ denotes three Euler angles and three translation parameters, and $W^{-1}$ is the inverted warp. In this way, transformation parameters are updated as $ \bm w_{i+1}= \bm w_i \Delta \bm w_i$ after the $i$-th iteration and $W_{i+1}$ = $Exp(\bm w_{i+1})$.   

\textbf{On the self-supervised learning:} We adopt the idea of self-supervised learning to enhance the representation ability of the feature extraction module. We especially extract proper local descriptors by using the high-level relation among different local geometries, as in the \textit{second term} in Eq. \ref{5}. The \textit{second term} can be considered as a pretext learning task\cite{kolesnikov2019revisiting}. The design of such a pretext task is based on the critical observation that local distinctive geometric structures of point clouds are transformed consistently under the same rigid transformation (see Fig. \ref{motivation}).

In the following, to avoid the abuse of all local geometries, we will elaborate on the intuition of selecting local distinctive subset points.

\textbf{On the selection of representative subset points.} We employ two representative points related to the barycenter of the source point cloud $X$ \cite{liu2022robust}, denoted by ${\bm x}_{b}=\frac{1}{N}\sum_{i=1}^{N}({\bm x}_{i}),$
the farthest point from barycenter,
\begin{equation}
{\bm x}_{f}=\arg \max_{{\bm x}_{i}\in X} \min (\| {\bm x}_{i} - {\bm x}_{b}\|_{2} \; , \; c),
\end{equation}
and the closest point from barycenter,
\begin{equation}
{\bm x}_{c}=\arg \min_{{\bm x}_{i}\in X} \| {\bm x}_{i} - {\bm x}_{b}\|_{2}.
\end{equation}
We also adopt an outliers rejection scheme to remove isolated points far from the barycenter \cite{yang2020teaser}. For example, one should reject outliers if they exceed the threshold $c$. Then we choose point ${\bm x}_{f}$ and search its $N_{l}$ nearest neighbors based on Euclidean distance, forming a local set $X_{1}$. Similarly, subset $X_{2}$ is selected among the point ${\bm x}_{c}$. The effect of local size $N_{l}$ on registration performance will be discussed in ablation studies. 

\par The intuition of the selection strategy is that (1) We want to capture \textit{distinctive geometric structures} of the point cloud by two subsets points to enhance the representation ability of the feature extraction module. From the perspective of geometry, the farthest and the closest points are located in \textit{distinctive regions} \cite{eldar1997farthest, shamos1975closest}. (2) We need these two subsets to be \textit{independent and without overlap}. Therefore, they are more likely to be independent if they are far from each other.

Since discovering features sensitive to rotations and translation from unlabeled source and target point clouds is quite challenging, a single feature extraction module may not lead to sufficient representations. If global and local descriptors own rich transformation awareness, proper supervision of local consistency will be offered, thus creating a virtuous circle for optimization. On the contrary, the learning process may be trapped into local minima for the lack of transformation awareness of extracted representations. To avoid this issue, we introduce two auxiliary tasks to enhance the rotation awareness of global and local descriptors.

\subsection{Auxiliary Task I: Self-Reconstruction}

The feature extraction module aims to learn a transformation awareness descriptor, which needs to reflect the effects of rigid-body transformation, enabling the modified inverse compositional algorithm to recover the transformation parameters $W$. We choose the vanilla PointNet \cite{qi2017pointnet} as the feature extraction module $\psi$, which is composed of Multi-Layer Perceptron (MLP) and a max-pooling operation.

\begin{figure}[t]
	\centering
	\includegraphics[width=0.9\columnwidth, height = 3.5cm]{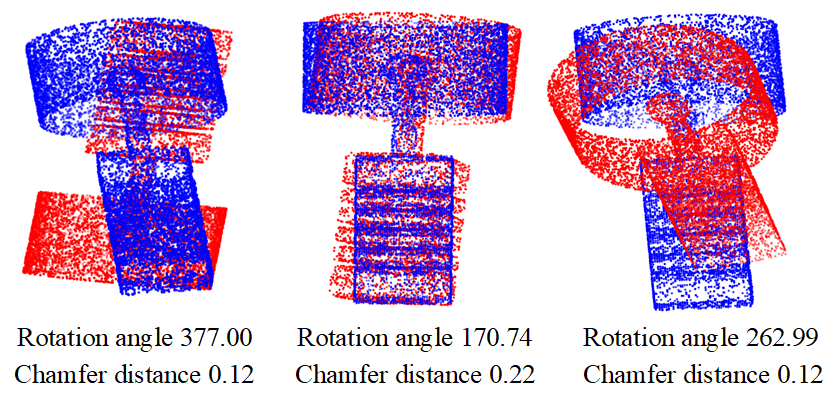}
	\caption{The weak rotation awareness of Chamfer Distance. The figure demonstrates Chamfer Distances between two point sets under rotation transformations. The unit of angular measurement is in degree.}
	\label{fig3}
\end{figure}

\begin{algorithm}
	\DontPrintSemicolon
	\KwIn{Source Point Cloud $X$ and Target Point Cloud $Y$; The number of epochs $N_{epoch}$ and the number of iterations $T$;}
	\KwOut{Rigid transformation $W$}
	Initialize the rigid transformation $W_0$, \\
	Initialize parameters of $\psi$, $\xi$, and $\ell$\\
	\While{not convergent }{
	Feature Extraction $\psi(X)$ and $\psi(Y)$ \\ 
	Compute self-reconstruction $\mathcal{L}_{\rm CD}$ and the normal estimation $\mathcal{L}_{\rm n}$ based on Eq. (14) and Eq. (15), respectively\\
	Freeze parameters of $\psi$\\
	Calculate $J \gets \frac{\partial \psi(W_0^{-1}(Y))}{\partial \bm w}$\\
	Initialize  $\bm w_0$, $i \gets 0, T \gets 10, threshold \gets  10^{-7}$\;
	\While{$i \leq T$ and $||\Delta \bm w_i||_2 > threshold$}{
		Compute $\Delta \bm w_i \gets ({J^{T}J})^{-1}J^{T} (\psi(W_{i}(X))-\psi(Y))$\\
		Update $\bm w_{i+1} = \bm w_{i} \Delta \bm w_i$\\
		$W_{i+1}$ = $Exp(\bm w_{i+1})$ \\
		$i \gets i + 1$\; 
	}
	Compute the primary loss $\mathcal{L}_{\rm p}$ based on Eq. (9)\\
	Update parameters of $\psi$, $\xi$, and $\ell$ using the primary loss $\mathcal{L}_{\rm p}$ and auxiliary losses $\mathcal{L}_{\rm CD}$ and $\mathcal{L}_{\rm n}$ based on Eq. (8)
}
	\caption{{\sc DVDs Algorithm}}
	\label{alg1}
\end{algorithm}

In the self-reconstruction task, we employ a folding-based \cite{yang2018foldingnet} decoder as the reconstruction module $\xi$. Besides, Chamfer Distance \cite{Fan2017chamfer} is adopted to define the reconstruction error, measuring differences between the original point cloud $P$ and the reconstructed version $P^{'}$:
\begin{equation}
{{\cal L}_{{\rm{CD}}}}\left( P \right) = \frac{1}{{\left| P \right|}}\sum\limits_{u \in P} {\mathop {\min }\limits_{v \in {P^{'}}} } \left\| {u - v} \right\| + \frac{1}{{\left| {{P^{'}}} \right|}}\sum\limits_{v \in {P^{'}}} {\mathop {\min }\limits_{u \in P} } \left\| {u - v} \right\|,
\end{equation}
This work will take both the source and the target points for reconstruction, \emph{i.e.}, $P \subset \left\{ {X,Y} \right\}$. As a result, we have
\begin{equation}
{{\cal L}_{\rm CD}} = \sum\nolimits_{P \subset \left\{ {X,Y} \right\}} {{{\cal L}_{\rm CD}}\left( P \right)}.
\label{AT1}
\end{equation}
However, Chamfer Distance pays more attention to the translation than the rotation and is blind to certain visual inferiority \cite{achlioptas2018learning}. Fig. \ref{fig3} demonstrates such an example. Two-point clouds with large rotation angles may lead to a small Chamfer Distance loss, resulting in weak rotation awareness. Therefore, we further consider another auxiliary task for improving the rotation awareness of global and local descriptors.

\subsection{Auxiliary Task II: Normal Estimation}

Normal estimation is an essential step in many research areas of point cloud \cite{liu2019relation}, \emph{e.g.}, rendering, and surface reconstruction. Unlike pursuing estimation precision, we aim to improve the rotation awareness of global and local descriptors by the normal estimation task. We use a lightweight MLP $\ell$ to generate the estimated normals with the concatenated coordinates $p_{i}$ and global features $\psi(P)$. The estimation error is measured by the cosine loss function as 
\begin{equation}
{{\cal L}_{{\rm{n}}}} = \sum\limits_{P \subset \left\{ {X,Y} \right\}} {\left( {1 - \frac{1}{N}\sum\limits_{i = 1}^N {\cos \left( {\ell \left( {\left[ {{p_i},\psi (P)} \right]} \right),p_i^{{\rm{normal}}}} \right)} } \right)},
\label{AT2}
\end{equation}
where $p_i^{{\rm{normal}}}$ denotes the ground truth normal of the point $p_i$.
Combing results in Eq. \eqref{5}, Eq. \eqref{AT1}, and Eq. \eqref{AT2}, we finally obtain the overall objective as shown in Eq. \eqref{PI1}. Besides, we have $\lambda_{1} = 0.5$ and $\lambda_{2}=0.1$, which will be further discussed in Section \ref{experiments}. Since we use the above Deep Versatile Descriptors to solve the point cloud registration, the proposed method is named DVDs. For readers' convenience, the flow of the proposed method is given in Algorithm \ref{alg1}, in which the convergence condition denotes that the maximum epoch number is reached or the validation loss does not decrease anymore in ten consecutive training epochs.

\begin{figure*}[t]
	\centering
	\includegraphics[width=0.9\columnwidth]{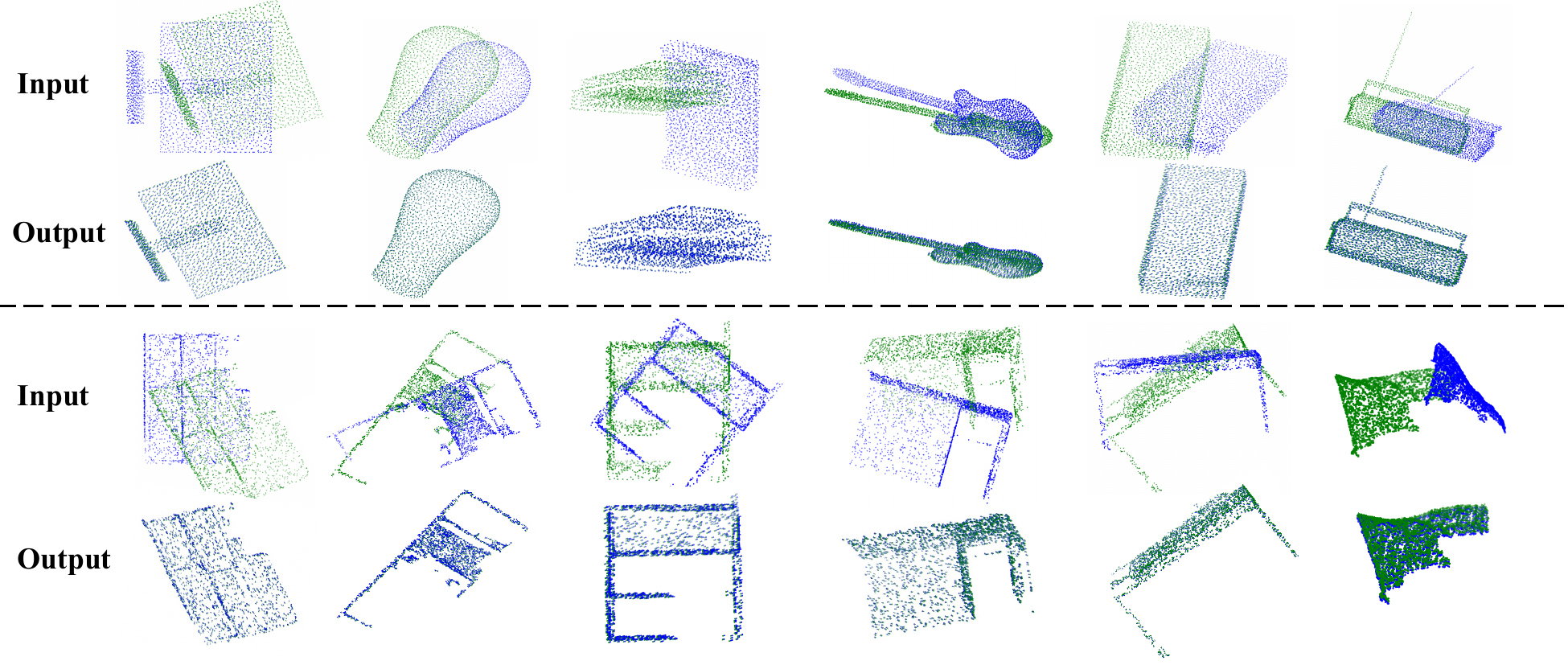} 
	\caption{Qualitative results of the proposed method on the ModelNet40 dataset and the ScanObjectNN dataset. \textbf{Input} and \textbf{Output} represent the proposed method's input and output, respectively. The top row demonstrates the results from ModelNet40, and the last row shows the results from ScanObjectNN.}
	\label{vis}
	\vspace{-0.3cm}
\end{figure*}

\begin{figure*}[t]
	\centering
	\includegraphics[width=0.95\columnwidth]{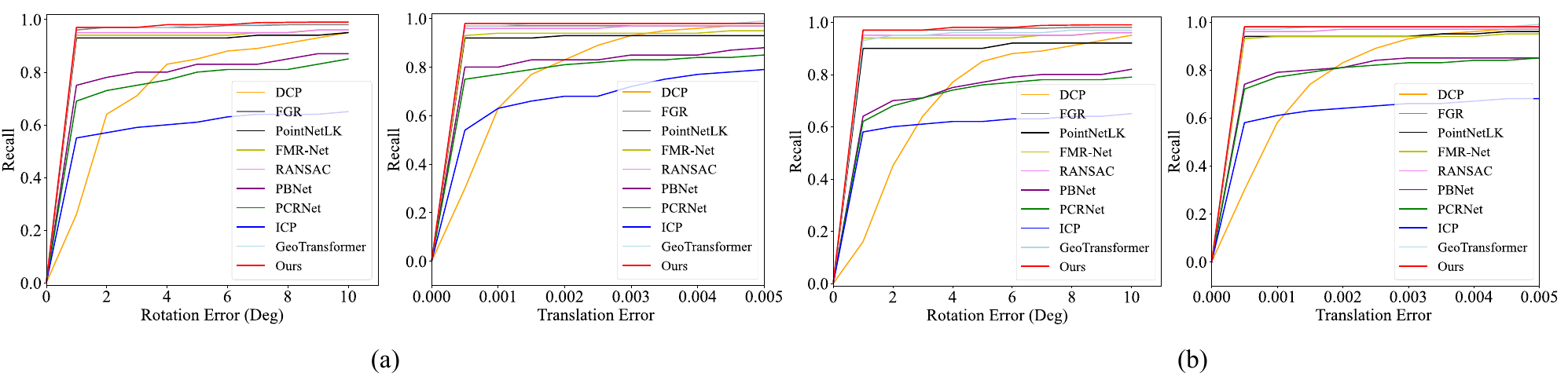} 
	\caption{Evaluation on the ModelNet40 dataset. All employed deep learning methods are trained on the training set of the first 20 categories in ModelNet40. (a) Evaluation over unseen objects. All these models are evaluated on the test set of the first 20 categories in ModelNet40. (b) Evaluation over unseen categories. All these models are evaluated on the test set of the last 20 categories in ModelNet40.}
	\label{fig5}
	\vspace{-0.3cm}
\end{figure*}

\section{Experiments}
\label{experiments}
Intelligent transportation systems inevitably contain many challenging conditions, such as noisy data, outliers, partial visibility, etc. As a result, this section conducts extensive experiments to verify the effectiveness of the proposed method. 

{\bf Experiments organization:} Firstly, we conduct effectiveness and robustness evaluations, including evaluations on synthetic object-centric point cloud datasets (Subsections \ref{unseen_object} and \ref{unseen_category}), cross dataset evaluation (Subsection \ref{cross}), real-world 3D indoor scenes (Subsection \ref{large}), and large-scale real-world outdoor scenes (Subsection \ref{large-kitti}). Moreover, we show the generality of the proposed local descriptor (Section \ref{general}).  Finally, the complexity analysis (Section \ref{efficiency}) and ablation studies (Section \ref{ablation}) are given to show the efficiency of DVDs and the effectiveness of the individual module in the novel DVDs, respectively. \\  

{\bf Experimental protocol:} We conducted comprehensive experiments on four datasets, including ModelNet40\cite{wu20153d}, ScanObjectNN\cite{uy2019revisiting}, 7scenes\cite{shotton2013scene}, and KITTI odometry\cite{geiger2012we} benchmarks. ModelNet40 is a synthetic dataset consisting of 12,311 CAD models. We randomly sample 1024 points from each CAD model 2 times with different random seeds and name sampled point clouds as the source point cloud $X$ and the target point cloud $Y$, respectively \cite{xu2021omnet}. We rescale point clouds into a unit sphere. Following experimental protocol in\cite{li2021pointnetlk}, all deep learning-based models are trained with the first 20 categories of the training set in ModelNet40. Euler angles are uniformly sampled in the range [0, $45^{\circ}$] and translations in the range $[-0.5, 0.5]$ for each axis during the training process \cite{wang2019deep, zhu2021point}. We transform the source point cloud $X$ through the sampled rigid transformation. The source point cloud $X$ and the target point cloud $Y$ are fed into the network, which aims to register the two point clouds. The maximum number of iterations $T$ is set as 10. We train the model for a total of 200 epochs. All experiments are implemented on a single NVIDIA Titan Xp. Note that during the evaluation process, the reconstruction task and the normal estimation task are not used.

\begin{figure*}[t]
    \centering
        \includegraphics[width = 0.95 \columnwidth]{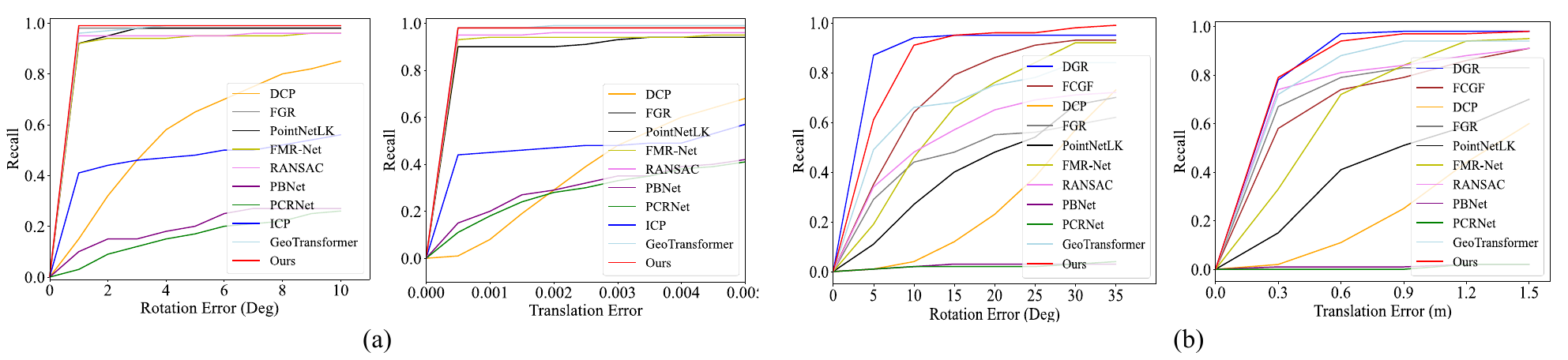}
        \caption{Evaluation on real-world datasets. All employed deep learning models are trained on the training set of the first 20 categories in ModelNet40. (a) Cross dataset evaluation on the test set of the real-world ScanobjectNN dataset. (b) Evaluation of the test set in the indoor 7-Scenes dataset.}
        \label{fig7}
        \vspace{-0.3cm}
\end{figure*}

\begin{figure*}[t]
	\centering
	\includegraphics[width=0.9\columnwidth]{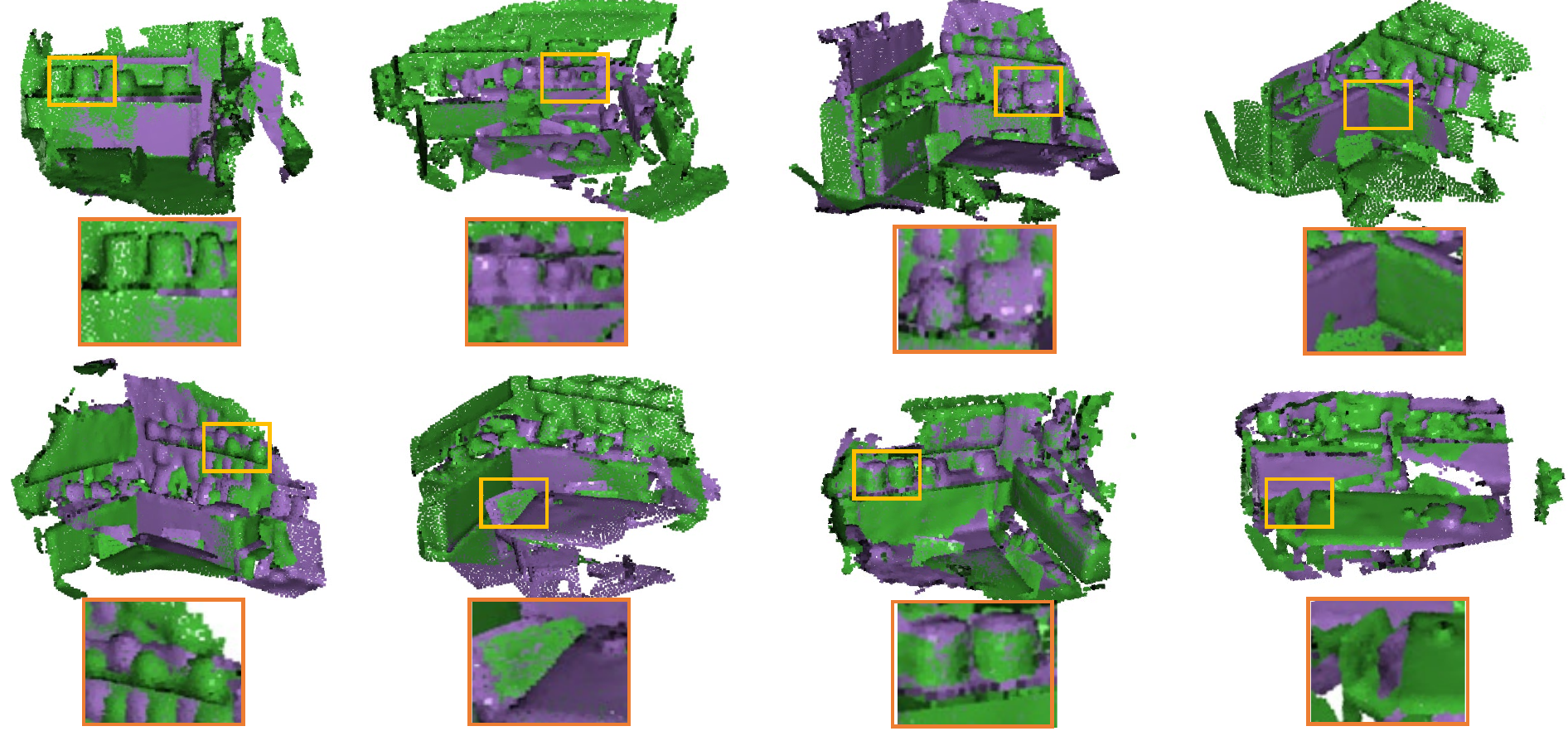} 
	\caption{Qualitative results of the proposed method on the 7Scene dataset. Purple represents the transformed point cloud, and green is the target point cloud.}
	\label{fig8}
	\vspace{-0.3cm}
\end{figure*} 

\subsection{Effectiveness and Robustness Evaluations}

\subsubsection{\textbf{Evaluation on Unseen Objects}}
\label{unseen_object}
In the first experiment, we compare our method with the following algorithms: ICP\cite{besl1992method}, PointNetLK\cite{aoki2019pointnetlk}, PCRNet\cite{sarode2019pcrnet}, FMR-Net\cite{huang2020feature}, DCP\cite{wang2019deep}, FGR\cite{zhou2016fast}, PBNet\cite{zheng2022global}, GeoTransformer\cite{qin2022geometric}, and RANSAC\cite{fischler1981random}. PBNet consists of a deep learning stage and a traditional optimal search stage. For a fair comparison, PBNet is evaluated with the deep learning stage. All models are evaluated on the test set of the first 20 categories in the ModelNet40 dataset. 

We first show the qualitative results in Fig. \ref{vis}. Differences between predicted values and ground truth are measured by the root mean squared error (RMSE). A successful registration is confirmed if rotation and translation errors are smaller than predefined thresholds \cite{choy2020deep}. Performances are evaluated by the recall, \textit{i.e.}, the percentage of successful registration point cloud pairs. 
Fig. \ref{fig5}(a) shows the comparison results of all employed methods. All deep learning-based registration methods outperform the classical ICP, proving the effectiveness of introducing deep learning technologies for the registration problem. The superior performance can be explained by the fact that deep learning-based registration algorithms utilize massive data and learn powerful representations. Moreover, the proposed method attains the best recall rate among all unsupervised registration ones and has comparable performance with the supervised algorithm GeoTransformer, proving the effectiveness of novel designs in the proposed method.

\subsubsection{\textbf{Evaluation on Unseen Categories}}
\label{unseen_category}
In this subsection, we evaluate the generalization of the deep learning methods on unseen categories. To this end, all methods are trained on the training set of the first 20 categories and then tested on the test set of the reminding 20 categories in the ModelNet40 dataset. Similar to the above Section, recall is used for the performance metric. Fig. \ref{fig5}(b) shows that PBNet and PCRNet are more susceptible to unseen categories than other deep learning methods. Furthermore, PointNetLK has a decreased performance as its encoder is difficult to extract useful representations from unseen categories. Lastly, the supervised method GeoTransformer, classical FGR, and the proposed method generalize well to unseen categories and are more robust than other deep learning-based approaches.

\subsubsection{\textbf{Generalization from ModelNet40 to the real-world ScanobjectNN}}
\label{cross}
This group of experiments evaluates the generalization ability of optimized deep learning models under the cross-dataset evaluation scheme, which is essential for practical applications in intelligent transportation systems. Following \cite{rao2020global, rao2022pointglr, li2021pointnetlk}, we train all deep learning models on the training set of the first 20 categories in ModelNet40 and then test them on the test set of ScanObjectNN \cite{uy2019revisiting}, which is a newly published real-world dataset containing 2902 objects with 15 classes. The results are reported in Fig. \ref{fig7}(a). Compared to case studies on the synthetic dataset, some deep learning methods, such as FMR-Net, PBNet, and PCRNet, show sharp performance degradation on the realistic dataset. On the other hand, the proposed method with transformation awareness designs is more robust than other methods, demonstrating the effectiveness of improving the transformation awareness of the feature extraction module.

\begin{figure*}[t]
	\centering
	\includegraphics[width=0.9\columnwidth]{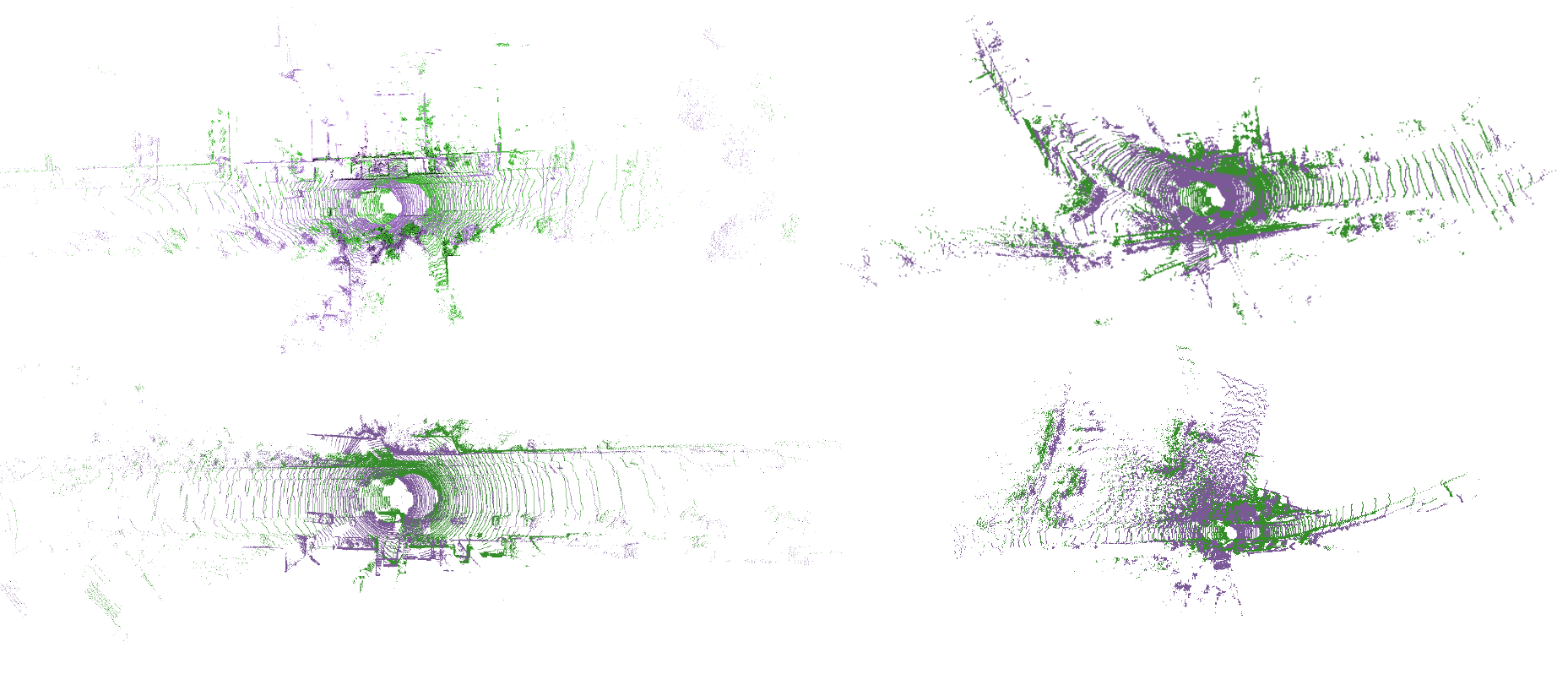} 
	\caption{Qualitative results of the proposed method on the KITTI dataset. Purple represents the transformed point cloud, and green is the target point cloud.}
	\label{fig:kitti_vis}
	\vspace{-0.5cm}
\end{figure*}

\subsubsection{\textbf{Generalization from ModelNet40 to Real-World indoor 3D Scenes}}
\label{large}

7Scene\cite{shotton2013scene} is a real-world indoor 3D Scenes registration benchmark. Following\cite{choy2020deep}, we generate 3D scan pairs with more than 70\% overlap and down-sample 2048 points for testing DCP, FMR-Net, PCRNet, PBNet, and PointNetLk. During the testing process, Euler angles are uniformly sampled in the range [0, $45^{\circ}$] and translations in the range [-0.5, 0.5]. In addition to the compared methods in the above subsection, we also compare with DGR \cite{choy2020deep} and FCGF \cite{choy2019fully}, which are the state-of-the-art methods on this challenging dataset. 
As DGR and FCGF could not be trained on the ModelNet40, and DGR requires a pre-trained FCGF model, we use all points for training DGR and FCGF on the 7scene. The voxelization strategy \cite{thomas2019kpconv, li2021pointnetlk, qin2022geometric} is adopted for processing large-scale point clouds\footnote{We use PointNet to obtain a high-dimensional vector as global descriptors of point clouds. Extensive experimental results on small-scale datasets (\emph{e.g.,} ModelNet40 and ScanObjectNN have shown their effectiveness. However, such global descriptors are challenged to represent complex and large-scale scenes. Therefore, we uniformly partition the large-scale point cloud into $K$ local voxels following the protocol shown in \cite{li2021pointnetlk}. During experiments, we set the number of voxels $K=27$ and points in each voxel keep their original accurate locations. Then, we use PointNet to extract a descriptor from each voxel. In this way, we capture the geometric structure of each voxel. Finally, we convert local voxel Jacobians to the global Jacobian and solve the point cloud registration. Note that such a voxelization strategy is only used during the testing process. In this work, we constantly use the voxelization strategy for large-scale point cloud processing.}. Fig. \ref{fig7}(b) shows that DGR and the proposed method outperform other deep learning ones. DGR achieves superior performance by using correspondence confidence prediction and weighted pose estimation, showing the effectiveness of high-level representations. On the other hand, the proposed method utilizes the global descriptor, local descriptors, and corresponding signal processing techniques to attain comparable performance with DGR, proving the validity of adopting the local descriptor. Lastly, Fig. \ref{fig8} shows the qualitative results of the proposed method.

\begin{figure}[t]
	\centering
	\includegraphics[width = 0.9 \columnwidth]{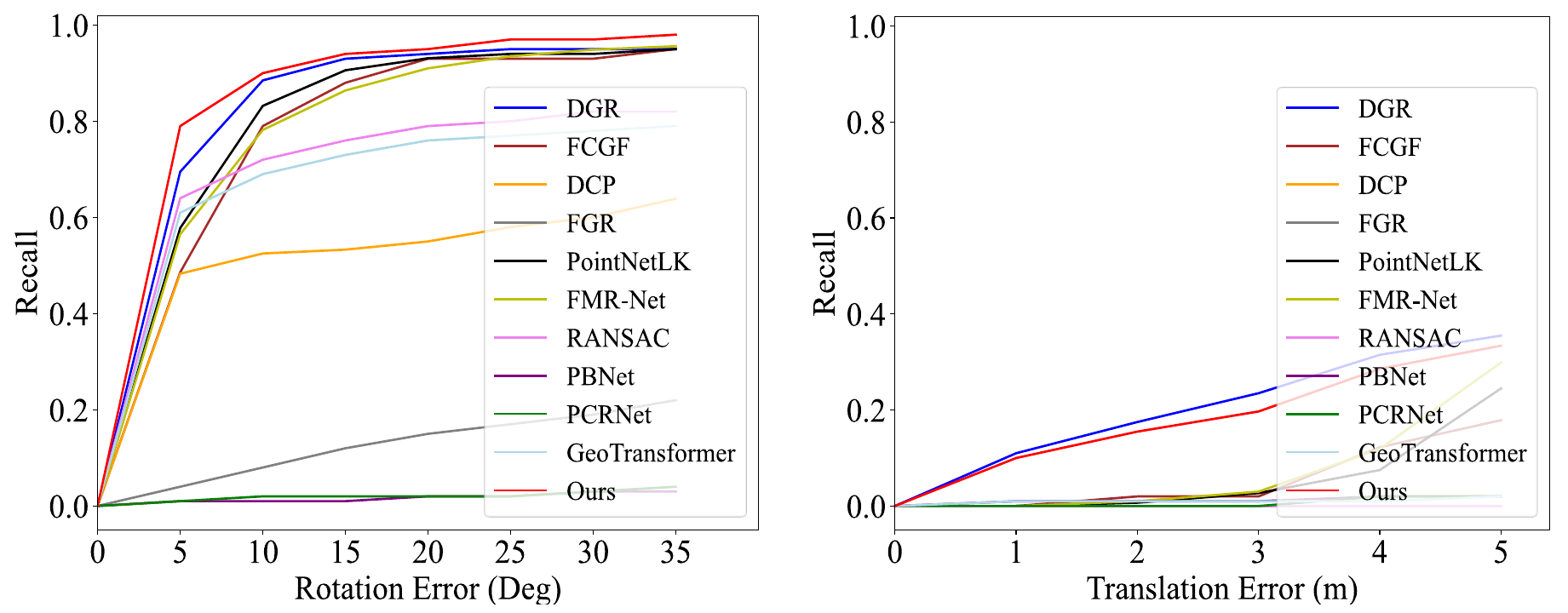}
	\caption{Evaluation on the KITTI odometry. Except for DGR and FCGF, all the deep learning-based models are trained on the training set of the first 20 categories in ModelNet40 and are evaluated on the KITTI dataset.}
	\label{fig:kitti}
	\vspace{-0.2cm}
\end{figure}

\begin{figure}[t]
	\centering
	\includegraphics[width = 0.95\columnwidth]{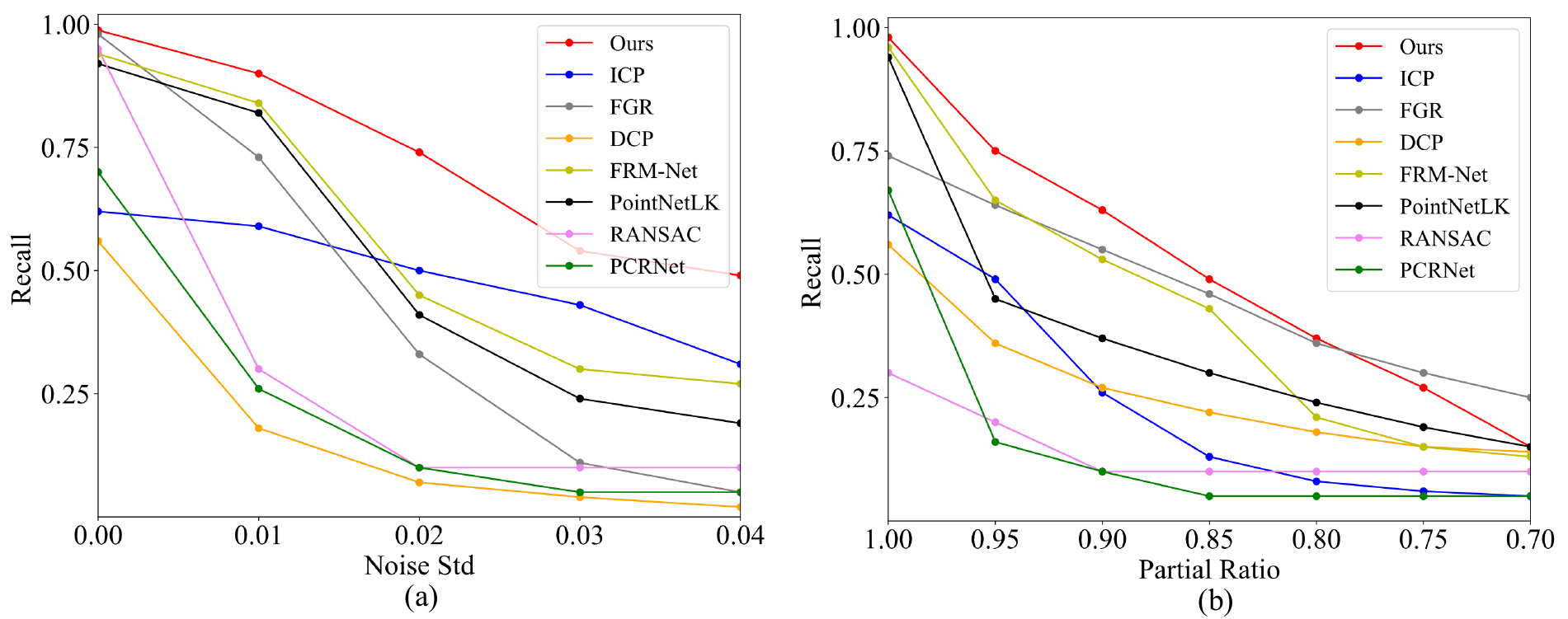}
	\caption{Robust Evaluation. All models are trained on the training set of the first 20 categories in ModelNet40 and are evaluated on the test set of the rest 20 categories in ModelNet40. (a) Robustness to noise. (b) Robustness to partial overlap.}
	\label{fig10}
\end{figure}

\subsubsection{\textbf{Generalization from ModelNet40 to Real-World outdoor 3D Scenes}}
\label{large-kitti}

KITTI odometry\cite{geiger2012we} is a real-world outdoor 3D Scenes benchmark containing 11 sequences of LiDAR-scanned driving scenarios. We follow\cite{qin2022geometric, choy2019fully} and use sequences 8-10 for testing. Only point cloud pairs more than 10 meters from each other are used for evaluation. Besides, we also refine the ground truth poses in the KITTI odometry with ICP\cite{qin2022geometric, choy2019fully}. Note that all the deep learning-based models are trained on the ModelNet20 except FCGF and DGR, which are trained on the 7scenes dataset. Qualitative results of the proposed method on the KITTI odometry dataset are shown in Fig. \ref{fig:kitti_vis}. Fig. \ref{fig:kitti} shows that although the proposed method is only trained on the small-scale synthetic dataset, it still achieves the best performance among all the compared methods on the rotation estimation. On the other hand, some deep learning models, including DCP, PCRNet, and PBNet, fail to handle large-scale outdoor point clouds because of the considerable difference between the distribution of the training set and the test set. In practice, processing data mismatching in industrial applications is challenging yet essential. As shown in Fig. \ref{fig:kitti}, the proposed method obtains predominant performance. Thus, it could be promising for autonomous driving tasks with numerous differently distributed point clouds from many scenarios.

\subsubsection{\textbf{Robustness Evaluation}}
\label{robustness}
In this section, we evaluate models in the presence of outliers and partial visibility. The success registration criteria are defined by a rotation error under $2^{\circ}$ and a translation error under 0.01. For the outliers scenario, noises are sampled from Gaussian distribution and are added to each point of source and target point cloud. Specifically, all the models are trained on clean data without adding noise augmentation. Though FGR and RANSAC methods are comparative with our method with noise-free data, they are sensitive to noisy data, as shown in Fig. \ref{fig10} (a). Adding noise to the source and target point cloud breaks the exact point-point correspondences. Therefore, the correspondence-based DCP shows limited performance. In contrast, the learning-based registration approaches without correspondences, such as FMR-Net, PointNetLK, and the proposed method, are more robust to noise.   
	
We also conduct experiments on partially visible data, as the acquired point clouds from intelligent vehicles are often partially due to occlusions. We generate partial source and target point clouds to simulate this condition independently from random camera poses following settings in \cite{aoki2019pointnetlk}. Note that all methods are trained on noise-free and fully-visible data from the first 20 categories in ModelNet40. Fig. \ref{fig10} (b) demonstrates that the proposed unsupervised model outperforms all the others against partial visibility because of utilizing both global and local descriptors. Therefore, it is promising to extend the proposed method to real-world intelligent transportation systems.

\setlength{\tabcolsep}{2pt} 
\begin{table}[t]
	\begin{center}
		\caption{Local consistency loss can be integrated into various point cloud registration models. LC denotes the proposed local consistency loss. All models are trained on the training set of the first 20 categories in ModelNet40 and are evaluated on the test set of the first 20 categories in ModelNet40.}
		\begin{tabular}{ccccc}
			\toprule
			Method  & RMSE($R$)  & MAE($R$)  & RMSE($t$) & MAE($t$)  \\
			\midrule
			DCP \cite{wang2019deep} &  4.834 & 2.795  & 0.020 & 0.010 \\
			PRNet \cite{wang2019prnet} & 1.506 & 0.874  & 0.015 & 0.010 \\
			IDAM \cite{li2020iterative} & 2.084  & 1.122 & 0.022 & 0.014\\
			DeepGMR \cite{yuan2020deepgmr} & 3.472 & 1.526  & 0.004 & 0.001 \\
			\hline
			LC + DCP & 3.950 & 2.298 & 0.020 & 0.010 \\
			LC + PRNet & 1.293 & 0.783 & 0.014 & 0.010\\
			LC + IDAM  & 1.527 & 0.933 & 0.016 & 0.010\\
			LC + DeepGMR & 3.192 &  1.247 & 0.004 & 0.001\\
			\bottomrule
		\end{tabular}
	\label{table4}
	\end{center}
\end{table}

\begin{table}[t]
\setlength{\tabcolsep}{11pt}
	\caption{Running time comparisons for registering a point cloud pair (in milliseconds). All models are evaluated on the test set of the ModelNet40.}\smallskip
	\centering
	\smallskip\begin{tabular}{cccc}
		\toprule
		Points & 500 & 1000 & 2000 \\
		\hline
		ICP & 7 &16 & 35\\
		PointNetLK & 70 & 73 & 75\\
		DCP & 6 & 9 & 20\\
		FGR & 143 & 172 & 232\\
		RANSAC & 18 & 48 & 110\\
		FMR-Net & 68 & 70 & 72\\
		PCRNet & 14 & 16 & 18\\
		Ours & 66 & 68 & 69\\
		\bottomrule
	\end{tabular}
	\label{table3}
	\vspace{-0.2cm}
\end{table}

\setlength{\tabcolsep}{2pt} 
\begin{table}[t]
	\begin{center}
		\caption{Ablation studies of the proposed model. The proposed method is evaluated on the test set of the first 20 categories in ModelNet40. If not specified, hyper-parameters in each model are manually optimized.}
		\begin{tabular}{ccccc}
			\toprule
			Model  & RMSE($R$)  & MAE($R$)  & RMSE($t$) & MAE($t$)  \\
			\midrule
			A &  4.201 & 0.525  & 0.023 & 0.002 \\
			B & 2.119 & 0.250  & 0.013 & 0.002 \\
			C  & 1.620 & 0.132 & 0.008 & 0.000\\
			D  & 1.095 & 0.101  & 0.007 & 0.000 \\
			\hline
			$|X_1|$ = 32 & 2.008 & 0.206 & 0.012 & 0.001 \\
			$|X_1|$ = 64 & 1.095 & 0.101  & 0.007 & 0.000 \\
			$|X_1|$ = 96  & 2.424 & 0.260 & 0.013 & 0.001\\
			\hline
			$\lambda_1=0.5, \lambda_2=0.5$ & 1.793 & 0.193 & 0.011 & 0.001 \\
			$\lambda_1=1, \lambda_2=0.1$ & 1.747 & 0.174  & 0.009 & 0.000 \\
			$\lambda_1=0.5, \lambda_2=0.1$  & 1.095 & 0.101  & 0.007 & 0.000\\
			\bottomrule
		\end{tabular}
		\label{ablation_all}
	\end{center}
\end{table}

\begin{figure*}[t]
	\centering
	\includegraphics[width=0.95\textwidth]{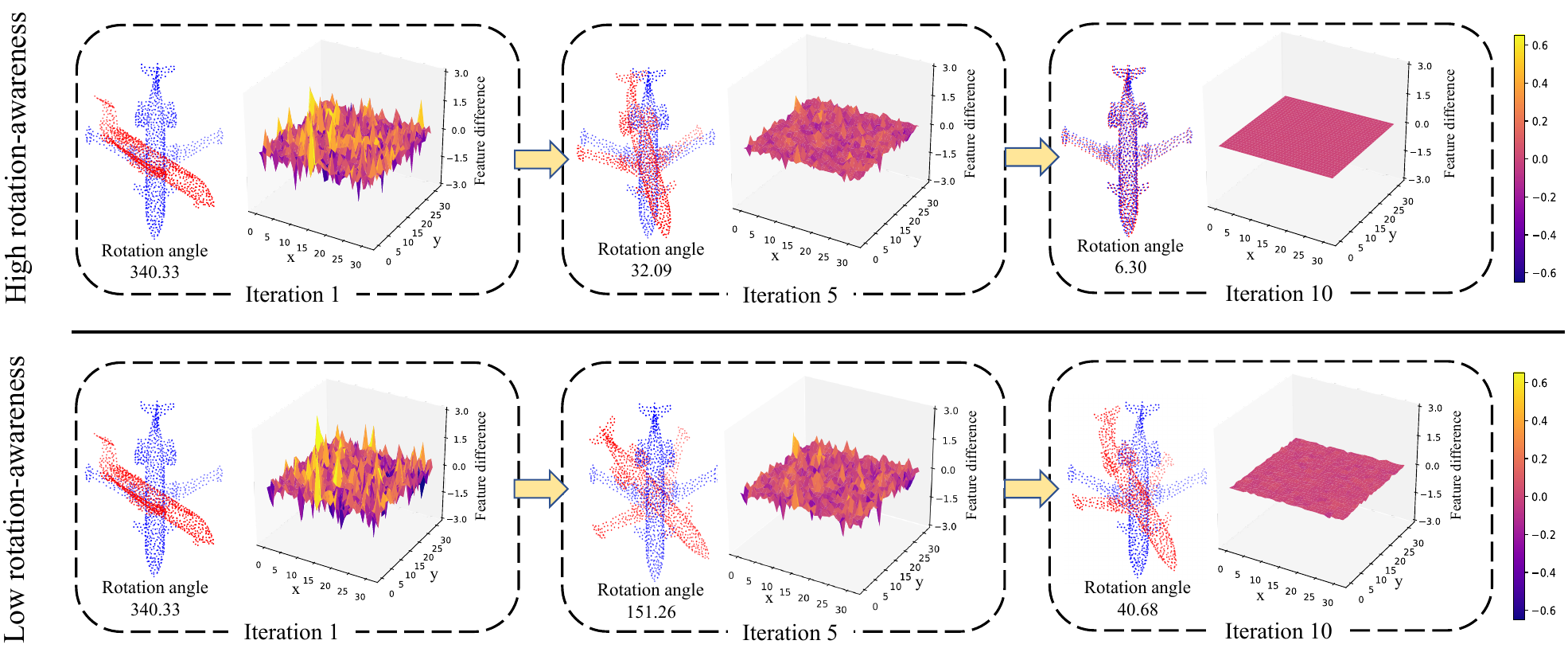} 
	\caption{We demonstrate the optimization process (different iterations) of the registration method with a high rotation-awareness feature extraction function (the top row) and a low rotation-awareness feature extraction function (the bottom row). Red and blue represent the transformed source point cloud $\hat X$ and the target point cloud $Y$, respectively. }  
	\label{fig4}
	\vspace{-0.3cm}
\end{figure*}

\setlength{\tabcolsep}{5pt} 
\begin{table*}[t]
	\begin{center}
		\caption{The comparison between DVDs and previous unsupervised registration methods. * denotes the proposed method.}
		\label{comparison}
		\begin{tabular}{cccccc}
			\toprule
			Method     & Unsupervised & Transformation awareness & Local descriptors & Computational complexity & Performance \\
			\hline
			DCP & \XSolidBrush   & \XSolidBrush  & \XSolidBrush & low & moderate  \\
			PointNetLK & \XSolidBrush  & \XSolidBrush & \XSolidBrush & moderate  & moderate \\
			PCRNet  & \checkmark   & \XSolidBrush  & \XSolidBrush & low & low \\
			FRM-Net  & \checkmark   & \XSolidBrush & \XSolidBrush & moderate  & moderate \\
			DVDs* & \checkmark   & \checkmark & \checkmark & moderate  & high\\
			\bottomrule
		\end{tabular}
	\end{center}
	\vspace{-0.3cm}
\end{table*}

\subsection{Generality of Local Consistency Loss}
\label{general}

We use the proposed local consistency loss to improve the performance of several classical point cloud registration models, including DCP \cite{wang2019deep}, PRNet \cite{wang2019prnet}, IDAM \cite{li2020iterative}, and DeepGMR \cite{yuan2020deepgmr}. All models are trained on the first 20 categories and tested on the rest 20 categories in ModelNet40. Tab. \ref{table4} indicates that being equipped with local consistency loss enables both the correspondence-based and correspondence-free registration methods to estimate more accurate transformation parameters. Specifically, using local consistency loss enables the rotation errors of DCP, PRNet, IDAM, and DeepGMR to decrease by about 10\% in terms of RMSE and MAE. Therefore, the proposed local consistency loss is a generic method complementary to the existing point cloud registration techniques.

\subsection{Computational Complexity Analysis}
\label{efficiency}

In this subsection, we evaluate the efficiency of the proposed method. The experiment has been conducted on the testing dataset of the last 20 categories from ModelNet40. Rigid transformations are randomly sampled with Euler angles in the range [0, $45^{\circ}$] and translations in the range $[-0.5, 0.5]$. Noises were sampled from Gaussian distribution and were added to each point of source and target point cloud. We performed this experiment on a 2.10GHz Intel E5-2620 and an NVIDIA Titan XP. Besides, the ICP, FGR, and RANSAC were executed on the CPU. The average computational complexity of all the compared methods was shown in Tab. \ref{table3}. Representative methods such as ICP, DCP, and PCRNet have high efficiency in the case of small-scale point-cloud data.
On the other hand, the proposed method is faster when processing large-scale point clouds (\emph{e.g.}, 2000) than RANSAC, FGR, FMR-Net, and PointNetLK.

\subsection{Rotation awareness Discussions}

Here, we provide an intuitive explanation of why the rotation awareness of the feature extraction function is essential for registering two-point clouds. Firstly, let us define the rotation awareness of the feature extraction function. Rotation awareness denotes the representation ability of the feature extraction function \emph{w.r.t.} rotation transformations. For example, the extracted feature of a point cloud is supposed to be changed when the point cloud is rotated. Specifically, given two point clouds with a large rotation angle, the high rotation-awareness feature extraction function should learn two different features from these two point clouds, \emph{i.e.,} the difference between two point clouds' features is vast. Inversely, the low rotation-awareness feature extraction function
may learn almost the same features from these two point clouds, \emph{i.e.,} the difference between two point clouds' features is negligible. Fig. \ref{fig4} visualizes feature differences of the transformed source point cloud $\hat X$ and the target point cloud $Y$ during the iterative optimization process. Moreover, we compare the registration process with a high rotation-awareness feature extraction function (in the top row) and a low rotation-awareness feature extraction function (in the bottom row). Finally, we reshape feature differences of the transformed source point cloud $\hat X$ and the target point cloud $Y$ into a square matrix for better visualization. 

Fig. \ref{fig4} shows that the feature difference decreases, and the registration becomes more accurate during optimization with the high rotation-awareness feature extraction function, \emph{i.e.,} the rotation angle is reduced from 340.33 to 6.30. However, the feature difference decreases for the low rotation-awareness feature extraction, but the registration is trapped in a local minimum, \emph{i.e.,} the rotation angle is still 40.68. Specifically, the rightmost column of Fig. \ref{fig4} demonstrates that the extracted global features of two-point clouds are almost the same when the two-point cloud are not aligned. Therefore, the rotation awareness of the feature extraction function helps identify whether two-point clouds are aligned.

\subsection{Ablation Studies}
\label{ablation}

To examine the effectiveness of each component in Eq. \eqref{PI1}, we conduct detailed ablation studies on the ModelNet40 dataset. Experimental results are shown in Tab. \ref{ablation_all}.  Model A denotes the baseline method, which is trained by the first term in Eq. \eqref{5}, \emph{i.e.,} only using global descriptors. Another baseline, denoted by model B, advances model A by integrating it with the local consistency loss (the second term in Eq. \eqref{5}). Furthermore, we can get model C by combining the primary and self-reconstruction tasks (Eq. \eqref{AT1}). Model C promotes model B, which convincingly verifies the effectiveness of the self-reconstruction task. Our full model D is then obtained by incorporating all tasks, and it outperforms the others, demonstrating that all individual tasks contribute to superior performance.

What's more, we study the effect of the size of local geometries in model D, \emph{i.e.}, the cardinalities of $|X_1|$ and $|X_2|$. Tab. \ref{ablation_all} shows that the model achieves the best performance with 64 points over that with large (96 points) or small (32 points) local size. Besides, we also investigate the effect of different hyper-parameters $\lambda_1$ and $\lambda_2$ in model D. Tab. \ref{ablation_all} shows that the proposed method achieves the best performance with $\lambda_1=0.5$ and $\lambda_2=0.1$. If we set a larger $\lambda_2$, \emph{e.g.,} 0.5, the normal estimation may be too difficult for the feature extraction module, and the performance decrease slightly. On the other hand, if $\lambda_1$ is too large, \emph{e.g.,} 1, the feature extraction module may only focus on the reconstruction task. Hence, we adopt $\lambda_1=0.5$ and $\lambda_2=0.1$ to balance these two auxiliary tasks.

Comparisons against all compared methods are summarized in Tab. \ref{comparison}. It is noted that PCRNet, FRM-Net, and the proposed method are unsupervised, while DCP and PointNetLK are supervised methods. Furthermore, all the compared methods do not consider local descriptors or the transformation awareness of the feature extraction network. As demonstrated by the experimental results on synthetic and real-world datasets, DVDs obtain the best registration performance against all compared methods with moderate computational complexity, showing the novelty of all designs in the proposed DVDs, especially the local descriptor.

\section{Conclusion}
This paper proposed a novel unsupervised representation for robust point cloud registration based on global and local descriptors. Besides, the transformation awareness of the proposed DVDs has been further enhanced by two auxiliary tasks (reconstruction and normal estimation). Numerical experiments revealed the superiority of the proposed registration algorithm: 1) It outperforms several unsupervised methods with lower computational complexity based on high transformation awareness descriptors. 
2) Owing to novel designs in the proposed DVDs, we obtain superior performance in one synthetic dataset and three real datasets over several challenging conditions, such as unseen objects/categories, cross/real-world datasets, and noise/outliers presence. 3) The proposed local consistency loss is flexible and can be used to improve most existing methods. 
4) Efficiency and robustness verification also prove the feasibility of the proposed method in real-world intelligent transportation systems.

Although the proposed method obtains promising results in indoor and outdoor point cloud registration tasks, there is room for improvement in large-scale point cloud processing. For example, the performance of the proposed descriptors will decrease if the large-scale point clouds contain extreme outliers and noises. As a result, future work will focus on improving the robustness.

\ifCLASSOPTIONcaptionsoff
\newpage
\fi

\bibliographystyle{IEEEtran}
\input{paper_arxiv.bbl}

\end{document}